# Recognition of Offline Handwritten *Devanagari* Numerals using *Regional Weighted Run Length* Features

Pawan Kumar Singh[1], Supratim Das, Ram Sarkar, Mita Nasipuri

Department of Computer Science and Engineering, Jadavpur University

188, Raja S.C. Mullick Road, Kolkata-700032, West Bengal, India

{pawansingh.ju, supratimdas21, raamsarkar, mitanasipuri}@gmail.com

[1]Corresponding author:{pawansingh.ju@gmail.com}

*Abstract*— **Recognition of handwritten *Roman* characters and numerals has been extensively studied in the last few decades and its accuracy reached to a satisfactory state. But the same cannot be said while talking about the *Devanagari* script which is one of most popular script in India. This paper proposes an efficient digit recognition system for handwritten *Devanagari* script. The system uses a novel 196-element *Regional Weighted Run Length* (RWRL) features for the recognition purpose. The methodology is tested using five conventional classifiers on 6000 handwritten digit samples. On applying 3-fold cross-validation scheme, the proposed system yields the highest recognition accuracy of 95.02% using Support Vector Machine (SVM) classifier.**

## I. INTRODUCTION

Handwritten numeral recognition is one of the most important research areas in pattern recognition domain from past few decades. It has several applications such as mail sorting using pin code, processing of bank cheques, form processing of different government job applications etc. [1] Handwriting digit recognition system can be defined as the ability of a computer to receive and interpret handwritten input digits from sources such as paper documents, photographs, touch-screens and other devices without the interference of human. The domain of handwritten digit recognition can be divided into following two types: (a) On-line handwritten digit recognition, and (b) Off-line handwritten digit recognition [1-2]. On-line handwriting recognition involves the automatic conversion of text as it is written on a special surface of electronic devices, where a sensor keeps the pen-tip movements as well as pen-up/pen-down transitions. On the other hand, in off-line handwritten digit recognition system, the input digit image may be sensed "off-line" from a piece of paper by optical scanning and the digits are finally classified on the basis of some discriminative features.

Various researches have been developed for *Roman* digit recognition systems due to the worldwide acceptance of *English* language. Research on digit recognition for other regional languages has started lately. Although India is a multilingual country, however, most of the official work is done using only *Hindi* and *English*. This is due to the fact that *Hindi* is the most popular language in terms of total number of native speakers in India. With around 366 million speakers, this language also ranks third in the world [3]. *Devanagari* script is used for writing *Hindi* language. Not only *Hindi*, *Devanagari* script is used by over 120 languages in the world. Apart from *Hindi*, some popular languages which use *Devanagari* script are *Nepali, Pali, Marathi, Konkani, Sanskrit* etc. Most of ancient and modern Indian literature is documented using *Devanagari* script, thereby digitization and restoration of ancient documents requires efficient recognition techniques. Due to diverse handwriting styles of human beings, the digit images can vary in size, thickness, orientations, and shape dimensions. These variations make the recognition task much more challenging. Sample handwritten digits written in *Devanagari* script are shown in Fig. 1.

The paper is organized as follows: Section 2 presents a brief review of some of the previous approaches to recognition of handwritten *Devanagari* numerals whereas Section 3 illustrates the preparation of database and pre-processing required before the feature extraction. Section 4 introduces our proposed MOD features for handwritten *Devanagari* digit recognition system. Section 5 describes the performance of the system on realistic databases of handwritten digits and finally, Section 6 concludes the paper.

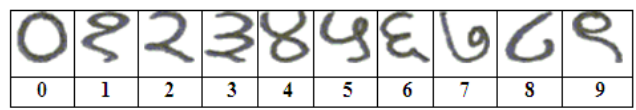

Fig. 1. Samples of digit images written in *Devanagari* script with their corresponding *Roman* numerals (printed below).

## II. RELATED WORKS

This section reviews some of the state-of-the-works related to *Devanagari* numerals. M. Hanmandlu *et al.* [4] proposed a Fuzzy model based recognition of handwritten *Hindi* numerals and obtained 92.67% accuracy. U. Bhattacharaya *et al.* [5]

proposed a Multi Layer Perceptron (MLP) neural network based classification approach for handwritten *Devanagari* numerals and obtained 91.28 % results. They used multi-resolution features based on wavelet transforms. A set of 17 geometric features based on pixel connectivity, lines, line directions, holes, image area, perimeter, eccentricity, solidity, orientation etc. were used for representing the numerals described by V. J. Dongre *et al.* [6]. Then, five discriminant functions *viz.*, Linear, Quadratic, Diaglinear, Diagquadratic and Mahalanobis distance were used for the classification purpose. Experimental results showed that Linear, Quadratic and Mahalanobis discriminant functions provided better results. Results of these three discriminants were then fed to a majority voting type combination classifier which showed 81.67% accuracy on a dataset of 1500 handwritten *Devanagari* digits. Three different kinds of features *namely*, density features, moment features and descriptive component features were extracted for classification of *Devanagari* numerals by R. Bajaj *et al.* [7]. They also proposed multi-classifier connectionist architecture for increasing the recognition reliability and obtained 89.6% accuracy. S. Arora *et al.* [8] proposed a two stage classification technique for the identification of *Devanagari* digits using some structural and statistical features. Here, zone based directional features, shadow based features, view based features and zone based centroid features were extracted. The classification was done using MLP classifier whereas the misclassified data were classified using Support Vector Machine (SVM) classifier. This method produced 93.15% recognition rate considering 18,300 digit images. U. Bhattacharya *et al.* [9] presented a two-stage classification system for recognition of handwritten *Devanagari* numerals. A shape feature vector computed from certain directional-view-based strokes of an input character image was used by both the Hidden Markov Model and Artificial Neural Network (ANN) classifiers of the present recognition system. The two sets of posterior probabilities obtained from the outputs of the above two classifiers were combined by using another ANN classifier. Finally, the digit images were classified according to the maximum score provided by the ANN at the second stage. In the proposed scheme, they achieved 92.83% recognition accuracy. G. Y. Tawde [10] presented a method of recognition of isolated offline handwritten *Devanagari* numerals using wavelets and NN classifier. The input digits were subjected to a single level wavelet decomposition using Daubechies-4 wavelet filter and the features were extracted considering only the LL band components. The MLP classifier showed an accuracy rate of about 70% for handwritten *Devanagari* numerals. V. J. Dongre *et al.* [11] used structural and geometric features to represent the *Devanagari* numerals. Each image was divided into 9 blocks and 8 structural features was extracted from each block. Similarly, 9 global features were also extracted from each of the entire digit images. These 81 features were classified using MLP classifier and 93.17% recognition accuracy was achieved on a total of 3000 handwritten samples of *Devanagari* numerals using 40 hidden neurons.

Research had so far concentrated on techniques that use more and more complicated feature detection and recognition methods. From the above discussion, it is clear that there are quite a few methods available for recognition of handwritten *Devanagari* numerals. However, their recognition rates are still far from the acceptable levels. Thus, work on recognition of handwritten *Devanagari* numerals is of considerable practical importance. In this paper, we propose a simple yet efficient feature extraction technique named *Regional Weighted Run Length* (RWRL) features for handwritten *Devanagari* numerals. Fig. 2 shows the block diagram of our proposed methodology.

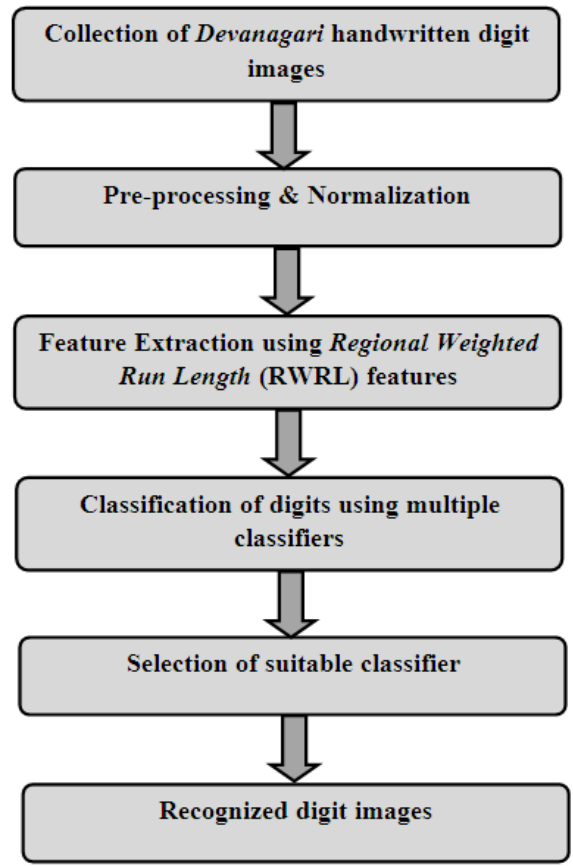

Fig. 2. Schematic diagram representing the key modules of our proposed methodology.

## III. DATASET PREPARATION AND PRE-PROCESSING

The generality of the results from any OCR endeavor depends a lot on the sample database used. The objective in data collection is to obtain a set of handwritten samples that capture effectively the possible variations in handwritings. This is necessary for getting a realistic assessment of any feature extraction methodology. Keeping in mind the above purpose, the following criteria were specified for collection of digit samples:

- The persons writing the digits were free to use different quality pens.

- They were required to write the digits in the specified grids while ensuring that the numerals did not touch the grid lines.

- Each person was asked to write 0-9 (in *Devanagari* script) five times in each column.

A database consisting of 6,000 samples were collected using the above approach. The sample datasheets were scanned through a HP scanner in gray scale format with 300 dpi resolution and stored in .bmp file format. A sample datasheet collected for the experiment is shown in Fig. 3. The digit images are cropped automatically from these scanned sheets to prepare the database for the experiment. These digit samples are enclosed in a minimum bounding square and are normalized to 64x64 pixels dimension. Gaussian filter [12] is used to remove any noisy pixels present in the handwritten digit images.

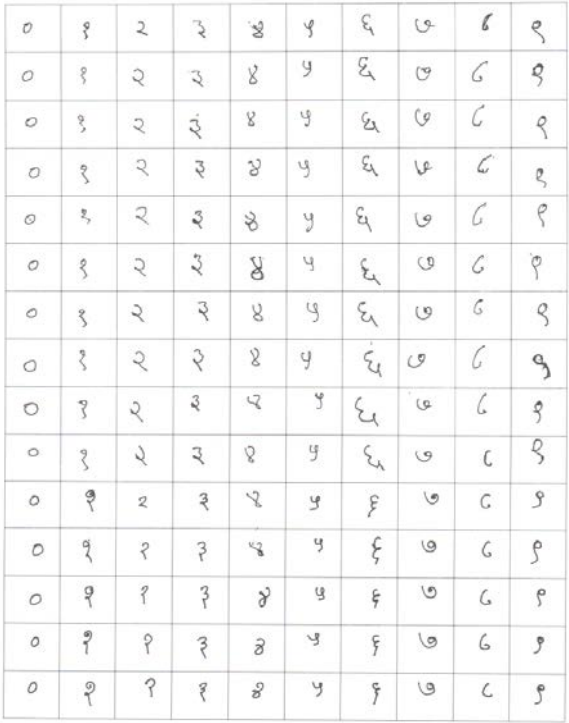

Fig. 3. A sample datasheet of handwritten *Devanagari* numerals used for our data collection purpose (the digits are written in numerical order from left to right).

## IV. *Regional Weighted Run Length* (RWRL) Feature Extraction

The main contribution of our work is the development of a new feature extraction technique which is named as *Regional Weighted Run Length* (RWRL) features. This feature extraction procedure includes the three key steps discussed below in detail:

### A. Contour Extraction

The pre-processing is followed by the contour extraction procedure. The contour of the digit images is computed and the feature vector is extracted from the pixels of contour. This is one of the advantages of the proposed work. Because considering only the contour pixels in feature calculation minimizes the overall computation time of the system.

Another point is that our purpose is to capture the shape variation of digit samples and which can be easily done by observing the contour of the same. The contour images of handwritten *Devanagari* numerals are shown in Fig. 4.

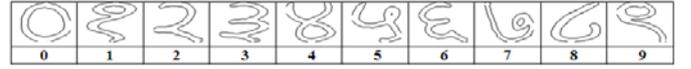

Fig. 4. Illustration of contour images of handwritten *Devanagari* numerals.

### B. Mask Orientation

Four different types of masks are employed depending on the orientation it analyses the data pixels such as *vertical*, *horizontal*, and the two oblique lines slanted at $\pm45^0$ to each black pixel. These four types of masks are illustrated in Fig. 5(a-d). Here, 8-connectivity neighborhood analysis is used to determine the direction of a black pixel.

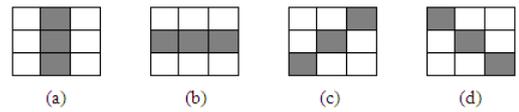

Fig. 5. Four different types of masks: (a) vertical, (b) horizontal, (c) $+45^0$ and (d) $-45^0$ used in the present work.

### C. Feature Extraction

The final step involves the extraction of features using our RWRL methodology. Here, the input digit image of size 64x64 is firstly divided into imaginary grids of size 8x8. Now, a mask of size 16x16 is made to shift in such a manner that each mask overlaps the preceding 8 pixels (both in horizontal and vertical directions) of the adjacent grid (for detail, refer to Fig. 6). After that, four concentric regions (squares) are formed considering each such mask *namely*, $R_i$, i=1,2,3,4. $R_1$ covers a region of size 4x4 in the center of the corresponding sub-image. $R_2$ comprises a region of size 8x8 excluding region $R_1$. $R_3$ is selected as a region of size 12x12 which excludes the preceding two regions. Finally, $R_4$ is the region of 16x16 dimensions excluding all three prior regions. This concentric region formation is illustrated in Fig. 7. In order to give importance to the local information, we have allocated certain weightages to the different run lengths estimated along four masks, as said earlier, at various regions $R_i$. It is visually noticed that the amount of discrimination for digit patterns diminish starting from the inner region to the outer region. That is why the center region, $R_1$, is given the highest weightage and $R_2$ is given the second highest weightage and so on. The estimated value of run length at a particular mask is measured using the formula given in Eq. (1).

$$F = \sum_{i=1}^{n} 2^{n-i} * R_i \qquad (1)$$

where, $F$ is the feature value calculated at regions $R_i$, i=1, 2, 3, 4 and n is the number of regions chosen for the present

TABLE I.    STATISTICAL PERFORMANCE ANALYSIS OF INDIVIDUAL HANDWRITTEN *DEVANAGARI* NUMERALS ACHIEVED BY SVM CLASSIFIER.

| Numeral | Statistical Performance Measures | | | | | | | | | |
|---------|--------------------|------|------|------|------|-----------|--------|-----------|------|------|
|         | Kappa Statistics | MAE | RMSE | TPR | FPR | Precision | Recall | F-measure | MCC | AUC |
| '0' |  |  |  | 0.985 | 0.001 | 0.987 | 0.985 | 0.986 | 0.984 | 0.992 |
| '1' |  |  |  | 0.973 | 0.006 | 0.951 | 0.973 | 0.962 | 0.958 | 0.984 |
| '2' |  |  |  | 0.947 | 0.005 | 0.951 | 0.947 | 0.949 | 0.943 | 0.971 |
| '3' |  |  |  | 0.933 | 0.007 | 0.935 | 0.933 | 0.934 | 0.927 | 0.963 |
| '4' | 0.9446 | 0.01 | 0.0998 | 0.973 | 0.006 | 0.945 | 0.973 | 0.959 | 0.954 | 0.984 |
| '5' |  |  |  | 0.938 | 0.006 | 0.943 | 0.938 | 0.941 | 0.934 | 0.966 |
| '6' |  |  |  | 0.963 | 0.004 | 0.962 | 0.963 | 0.963 | 0.958 | 0.980 |
| '7' |  |  |  | 0.940 | 0.005 | 0.953 | 0.940 | 0.946 | 0.940 | 0.967 |
| '8' |  |  |  | 0.932 | 0.006 | 0.949 | 0.932 | 0.940 | 0.934 | 0.963 |
| '9' |  |  |  | 0.917 | 0.008 | 0.926 | 0.917 | 0.921 | 0.913 | 0.954 |
| **Average** | **0.9446** | **0.01** | **0.0998** | **0.950** | **0.006** | **0.950** | **0.950** | **0.950** | **0.945** | **0.972** |

*MAE: Mean Absolute Error, RMSE: Root Mean Square Error, TPR: True Positive Rate, FPR: False Positive Rate, MCC: Matthews Correlation Coefficient, AUC: Area Under ROC.

case. Since four different types of masks are applied to each digit images, the feature vector for each digit images consists of 196 (49*4) dimension.

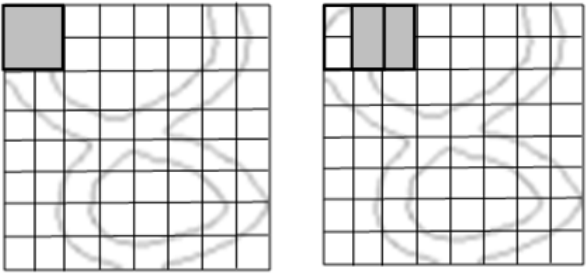

Fig. 6.    Illustration of choice of two consecutive masks for a sample handwritten *Devanagari* digit '4'.

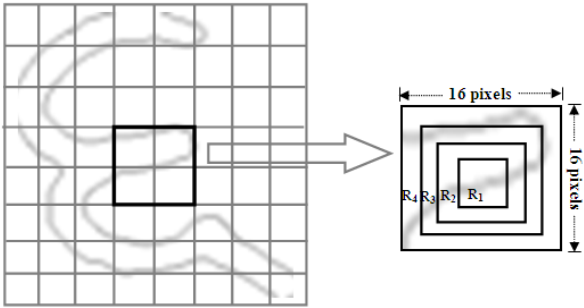

Fig. 7.    Example showing consideration of four regions on one sub-image of a sample handwritten *Devanagari* digit '6'.

## V.    PERFORMANCE EVALUATION

The proposed RWRL feature set has been applied on the prepared database of 6000 handwritten samples of *Devanagari* digits. For the experimentation, 4000 training samples are chosen by selecting 400 samples for each of 10 digit classes. A testing set of 2000 remaining samples are used by considering equal number of digit samples from each class. The experiments are carried out in Windows 7 operating system with Dual Core Processor 2.30 GHz and 2 GB RAM. MATLAB 2010a is the tool used for implementation. A set of five conventional classifiers have been applied for the classification purpose such as Naïve Bayes, MLP, Logistic, SVM and Random Forest. The recognition accuracies and their corresponding 95% confidence score for each of these classifiers are depicted in Fig. 8. From the figure it can be noted that the highest recognition accuracy is found to be 95.02% by SVM classifier using polynomial kernel. The detailed recognition results of individual numerals attained by SVM classifier are provided in Table I. From the experiment we have noted the maximum accuracy of 98.5% is achieved for *Devanagari* numeral '0'. The next highest accuracy of about 97.3% is achieved for both the numerals '2' and '4'. The lowest accuracy of 91.7% is observed for numeral '9'. To get the idea of the misclassified digits, the confusion matrix so produced by the SVM is also given in Table II. It is evident from this matrix that maximum misclassification error is observed between the digits '3' and '5'. The second most misclassification is seen where both the digits '7' and '8' get confused with the digit '9' and a few number of samples of '9' are found to be confused with '4'. The reasons for misclassification may be due to similarity of shape found between some of the digits, or sometimes different writing styles make the samples of a particular class closer to other class.

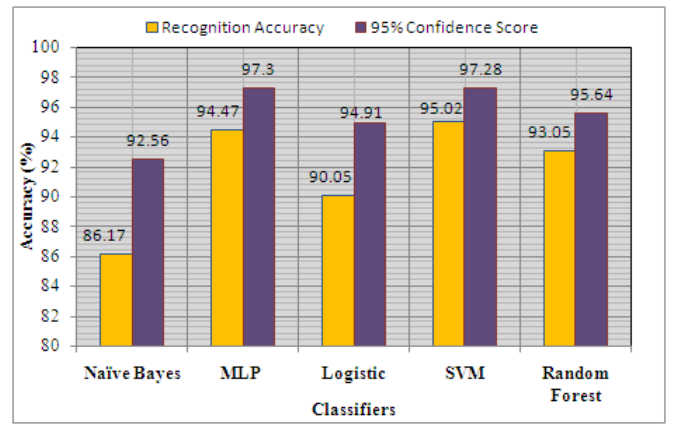

Fig. 8.    Graph showing the recognition accuracies and their corresponding confidence scores of our proposed methodology using five multiple classifiers.

TABLE II.  CONFUSION MATRIX FOR HANDWRITTEN *DEVANAGARI* NUMERALS GENERATED BY SVM CLASSIFIER.

| Numeral | Recognized as | | | | | | | | | |
|---|---|---|---|---|---|---|---|---|---|---|
| | '0' | '1' | '2' | '3' | '4' | '5' | '6' | '7' | '8' | '9' |
| '0' | **591** | 1 | 0 | 0 | 0 | 1 | 2 | 1 | 2 | 2 |
| '1' | 0 | **584** | 4 | 1 | 2 | 1 | 4 | 3 | 0 | 1 |
| '2' | 2 | 7 | **568** | 4 | 3 | 0 | 1 | 6 | 6 | 3 |
| '3' | 0 | 1 | 8 | **560** | 0 | 16 | 2 | 4 | 7 | 2 |
| '4' | 0 | 1 | 2 | 0 | **584** | 0 | 4 | 2 | 0 | 7 |
| '5' | 1 | 3 | 0 | 17 | 0 | **563** | 7 | 0 | 4 | 5 |
| '6' | 2 | 5 | 3 | 2 | 3 | 4 | **578** | 0 | 3 | 0 |
| '7' | 0 | 6 | 6 | 3 | 7 | 0 | 0 | **564** | 2 | 12 |
| '8' | 2 | 2 | 5 | 7 | 3 | 4 | 2 | 4 | **559** | 12 |
| '9' | 1 | 4 | 1 | 5 | 16 | 8 | 1 | 8 | 6 | **550** |

In the present work, we have also compared our proposed methodology with handwritten *Devanagari* numeral recognition methods proposed in the literature. This is illustrated in Table III. It can be noticed from Table III that our proposed methodology outperforms all the preceding numeral recognition techniques for handwritten *Devanagari* numerals.

## VI.  CONCLUSION

There are not much proficient works done towards the handwritten numeral recognition of *Indic* scripts. Among the *Indic* scripts, *Devanagari* script is the most popular script used in the Indian sub-continent. In this work, we propose a new RWRL features for the recognition of handwritten *Devanagari* numerals. A 196-dimension feature vector has been designed for the classification purpose. We have also created our own database containing 6000 samples of handwritten *Devanagari* numerals and a maximum accuracy rate of 95.02% is attained by SVM classifier using 3-fold cross validation scheme which is quite notable. The most important advantage of this feature extraction methodology is that it is less computational expensive than other state-of-the-art algorithms. The future aim will be to improve the present recognition accuracy by focusing more on the misclassification errors. The present methodology may also be integrated with some global complementary features in order to separate the confusing numerals. We will plan to increase the size of our current database so that it can be used as benchmarking dataset for the researchers to compare and validate their proposed algorithms on a common platform. We will also try to collect handwritten numerals written in other *Indic* scripts such as *Gujarati*, *Oriya*, *Gurumukhi*, *Kannada*, etc. The proposed method can then be implemented on these *Indic* scripts also which will confirm the generalization of our current work.

## ACKNOWLEDGEMENT

The authors are thankful to the Center for Microprocessor Application for Training Education and Research (*CMATER*) and Project on Storage Retrieval and Understanding of Video for Multimedia (SRUVM) of Computer Science and Engineering Department, Jadavpur University, for providing infrastructure facilities during progress of the work. The current work, reported here, has been partially funded by University with Potential for Excellence (UPE), Phase-II, UGC, Government of India.

## REFERENCES

[1] C.L. Liu, K. Nakashima, H. Sako, H. Fujisawa, "*Handwritten digit recognition: benchmarking of state-of-the-art techniques*", In: Pattern Recognition, vol. 36, pp. 2271-2285, 2003.

[2] M. Revow, C. K.I. Williams, G. E. Hinton, "*Using Generative Models for Handwritten Digit Recognition*", In: IEEE Transactions on Pattern Analysis and Machine Intelligence, vol. 18, no. 6, June 1996.

[3] http://web.archive.org/web/20071203134724/http://encarta.msn.com/media_701500404/Languages_Spoken_by_More_Than_10_Million_People.htmlRetrieved 2016-07-03.

[4] M. Hanmandlu, O.V. Ramana Murthy, ''*Fuzzy Model Based Recognition of Handwritten Hindi Numerals*'', In: Proc. of International Conference on Cognition and Recognition, pp. 490-496, 2005.

[5] U. Bhattacharya, B. B. Chaudhuri, R. Ghosh, M. Ghosh, "*On Recognition of Handwritten Devnagari Numerals*", In Proc. of the Workshop on Learning Algorithms for Pattern Recognition (in conjunction with the 18th Australian Joint Conference on Artificial Intelligence), Sydney, pp.1 -7, 2005.

[6] V. J. Dongre, V.H. Mankar, "*Devnagari Handwritten Numeral Recognition using Geometric Features and Statistical combination Classifier*", In: International Journal on Computer Science and Engineering (IJCSE), Vol. 5, No. 10, Oct 2013.

[7] R. Bajaj, L. Dey, S. Chaudhury, "*Devnagari numeral recognition by combining decision of multiple connectionist classifiers*", In: Sadhana, Vol.27, part. 1, pp.-59-72, 2002.

[8] S. Arora, D. Bhattacharjee, M. Nasipuri, M. Kundu, D. K. Basu and L. Malik, "*Handwritten Devnagari Numeral Recognition using SVM & ANN*", In: International Journal of Computer Science & Emerging Technologies (IJCSET), Vol. 1, Issue 2, pp.40-46, August 2010.

[9] U. Bhattacharya, S.K. Parui, B. Shaw, K. Bhattacharya, "*Neural Combination of ANN and HMM for Handwritten Devanagari Numeral Recognition*", In: 10th International Workshop on Frontiers in Handwriting Recognition, La Baule (France), Suvisoft, 2006.

[10] G. Y. Tawde, "*Optical Character Recognition for Isolated Offline Handwritten Devanagari Numerals Using Wavelets*", In: International Journal of Engineering Research and Applications, Vol. 4, Issue 2, pp. 605-611, 2014.

[11] V. J. Dongre, V. H. Mankar, "*Devanagari offline handwritten numeral and character recognition using multiple features and neural network classifier*", In: Proc. of IEEE International Conference on Computing for Sustainable Global Development (INDIACom), pp. 425-431, 2005.

[12] R. C. Gonzalez, R. E. Woods, "*Digital Image Processing*", vol. I. Prentice-Hall, India (1992).

TABLE III.    COMPARATIVE STUDY OF THE PROPOSED METHODOLOGY WITH STATE-OF-THE-ART TECHNIQUES.

| Researchers | Name of database | Size of database | Feature set used | Classifiers | Recognition accuracy (%) |
|---|---|---|---|---|---|
| H. Hanmandlu *et al.* [4] | Own database | Not Known | Vector distances | Fuzzy model | 92.67 |
| U. Bhattacharaya *et al.* [5] | Own database | Not Known | Multi-resolution features based on wavelet transform | MLP | 91.28 |
| V. J. Dongre *et al.* [6] | Own database | 1500 | Geometric features | Majority voting | 81.67 |
| R. Bajaj *et al.* [7] | Own database | Not Known | Density features, moment features and descriptive component features | Multi-classifier connectionist architecture | 89.6 |
| S. Arora *et al.* [8] | Own database | 18,300 | Structural and statistical features | MLP and SVM | 93.15 |
| V. J. Dongre *et al.* [11] | Own database | 3,000 | Structural and geometric features | MLP | 93.17 |
| **Proposed Methodology** | **Own database** | **6,000** | **MOD features** | **SVM** | **95.02** |